\documentclass{article} 
\usepackage{collas2022_conference,times}
\usepackage{booktabs}

\usepackage{amsmath,amsfonts,bm}

\def\eqref#1{equation~\ref{#1}}

\def\1{\bm{1}}

\def\vs{{\bm{s}}}

\def\vv{{\bm{v}}}

\DeclareMathAlphabet{\mathsfit}{\encodingdefault}{\sfdefault}{m}{sl}
\SetMathAlphabet{\mathsfit}{bold}{\encodingdefault}{\sfdefault}{bx}{n}

\newcommand{\softmax}{\mathrm{softmax}}

\newcommand{\KL}{D_{\mathrm{KL}}}

\DeclareMathOperator*{\argmin}{arg\,min}

\usepackage{hyperref}
\usepackage{dsfont}

\usepackage{xcolor}
\hypersetup{
    colorlinks=true,
    linkcolor=red,
    filecolor=magenta,      
    urlcolor=blue,
    citecolor=purple,
    pdftitle={Overleaf Example},
    pdfpagemode=FullScreen,
    }

\usepackage{qiangstyle}
\newcommand{\fs}[1]{\footnotesize $\pm$#1}

\title{Continual Learning and Private Unlearning}

\author{Bo Liu, Qiang Liu, Peter Stone \\
Department of Computer Science\\
The University of Texas at Austin\\
\texttt{\{bliu,lqiang,pstone\}@cs.utexas.edu} \\
}

\newcommand{\rebuttal}[1]{{\textcolor{blue}{#1}}}

\collasfinalcopy 

\begin{document}

\maketitle

\begin{abstract}
As intelligent agents become autonomous over longer periods of time, they may eventually become lifelong counterparts to specific people.
If so, it may be common for a user to want the agent to master a task temporarily but later on to forget the task due to privacy concerns.
However enabling an agent to \emph{forget privately} what the user specified without degrading the rest of the learned knowledge is a challenging problem.
With the aim of addressing this challenge, this paper formalizes this continual learning and private unlearning (CLPU) problem.
The paper further introduces a straightforward but exactly private solution, CLPU-DER++, as the first step towards solving the CLPU problem, along with a set of carefully designed benchmark problems to evaluate the effectiveness of the proposed solution. The code is available at \url{https://github.com/Cranial-XIX/Continual-Learning-Private-Unlearning}.
\end{abstract}
\section{Introduction}
Continual learning (CL) studies how an intelligent agent can learn continually over a sequence of tasks. In particular, when the agent is learning a new task, it is generally assumed that it loses access to data from previous tasks. As a result, the goal of a successful CL algorithm is to  \emph{forget} as little as possible about previous tasks while maximally adapting past knowledge to help learn the new task.

As deep learning has become increasingly popular, it has become generally known that straightforwardly applying stochastic gradient descent (SGD) on deep architectures when learning over a sequence of tasks leads to the so-called \emph{catastrophic forgetting} phenomenon~\citep{french1999catastrophic}, i.e., the network forgets much of what it learned previously when learning new knowledge. Thus, much CL research has focused on developing methods to mitigate forgetting. However, forgetting is \emph{not} always bad. Besides the fact that graceful forgetting---the process of deliberately compressing useful knowledge or removing useless knowledge---can help abstract learned knowledge and leave more ``space" for learning new knowledge~\citep{bjork2019forgetting}, we posit that it may also become common for an agent to be required to completely remove any trace of having learned a specific task.


For example, consider a robot manufacturing company that produces service robots, whose system is continually updated by learning \emph{novel} skills on the data collected from its customers' daily lives. From time to time, the company may be asked to expunge previously learned behaviors and/or knowledge about specific tasks that are found to raise potential fairness~\citep{mehrabi2021survey}, privacy or security issues~\citep{bae2018security}. Looking further into the future, consider another situation in which a person is undergoing a medical treatment plan and requests that their service robot learns to assist with the treatment. However, after having recovered, when a friend is about to visit, the person may not want the robot to exhibit any evidence of their previous medical treatment. In this case, the person would like to be able to request that the robot privately remove all knowledge of the treatment plan without impairing other unrelated knowledge it may have acquired during (or before or after) the time of the treatment.
Both of the above situations indicate that as personalized models that lifelong learn with and about humans become commonplace,
it is important for these models to carefully unlearn knowledge when necessary. This leads to the problem of machine unlearning (MU)~\citep{cao2015towards,bourtoule2019machine}. But to the authors knowledge, MU has not yet been well studied in the continual learning setting where the underlying data distribution can shift over time.


Note that even though catastrophic forgetting often happens naturally with rich parametric models such as deep neural networks, it might not be sufficient because 1) the user may want the agent to unlearn immediately (instead of unlearning over time) and 2) the unlearning must happen privately, meaning that after forgetting, it must not be possible to retrieve any information pertaining to the task, or even detect that the task has been previously learned.

With this motivation in mind, in this work, we present a novel but general CL problem setting where for each task, besides providing the task data, the user additionally provides a learning instruction indicating whether they want the agent to learn and remember the task permanently, to temporarily learn the task such that later on it will either forget or permanently remember it, or to forget a certain task completely and privately. We call this novel problem \emph{continual learning and private unlearning} (CLPU). To the best of our knowledge, only one previous paper discusses a similar problem setting pertaining to selective forgetting in continual learning~\citep{shibata2021learning}. However, the problem in that paper is different from CLPU as it defines forgetting as maximally degrading the performance on a task. As discussed in Sec.~\ref{sec:experiment}, this requirement is not privacy-preserving and can potentially leak information (e.g.,\ that the task has been previously learned).

To address CLPU, we propose a straightforward but exact method, named CLPU-DER++, based on both the dynamic architecture approach~\citep[e.g.][]{rusu2016progressive} and the rehearsal approach~\citep[e.g.][]{robins1995catastrophic} from the CL literature. Furthermore, we design a set of benchmark tasks along with novel evaluation metrics for evaluating any CLPU methods. To summarize, our main contributions are:
\begin{itemize}
    \item Formulating the continual learning and private unlearning (CLPU) problem.
    \item Presenting an initial solution, CLPU-DER++, to CLPU that achieves exact unlearning, and demonstrating its effectiveness on a novel set of benchmarks designed for CLPU.
\end{itemize}

\begin{figure}
    \centering
    \includegraphics[width=\textwidth]{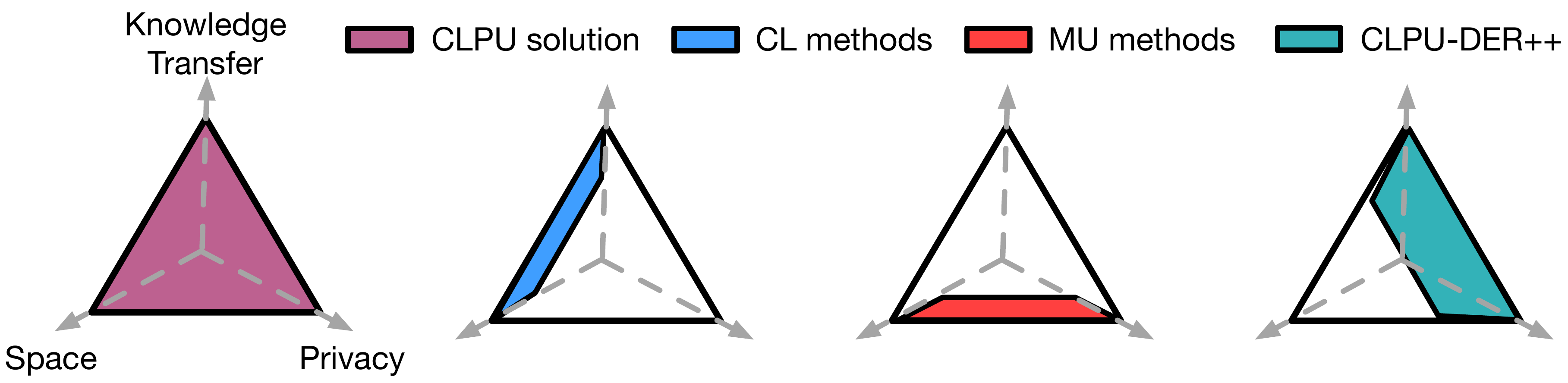}
    \caption{The CLPU problem has the Pareto front formed by good knowledge transfer ability, small model space, and no privacy leak. The ideal solution to CLPU achieves all of them simultaneously. We visualize what existing continual learning (CL) methods and machine unlearning (MU) methods achieve on the Pareto front above.  CLPU-DER++ represents an initial CLPU algorithm that achieves exact unlearning and good continual learning performance, in exchange for using model space.}
    \label{fig:pareto}
\end{figure}
\section{Related Work}
In this section, a brief review of continual learning and machine unlearning is provided. The relationship between CLPU and previous literature is summarized in Fig.~\ref{fig:pareto}.

\paragraph{Continual Learning} Continual learning (CL) assumes a learning agent learns continually over a sequence of tasks and in general the agent loses access to previous data when learning new tasks. Due to its generality, CL has been applied to a variety of areas including computer vision~\citep[e.g.][]{kirkpatrick2017overcoming}, reinforcement learning~\citep[e.g.][]{kirkpatrick2017overcoming,riemer2018learning}, natural language processing~\citep[e.g.][]{biesialska2020continual}, and robotics~\citep[e.g.][]{liu2021lifelong}. There exist three main approaches towards continual learning. 1) Dynamic architecture approaches study how to carefully and gradually expand the learning model to incorporate the learning of new knowledge~\citep{rusu2016progressive,yoon2017lifelong,mallya2018piggyback,rosenfeld2018incremental,mallya2018packnet,hung2019increasingly,hung2019compacting,wu2020firefly}. 2) Regularization-based methods design a regularization objective that prevents the model parameter deviating too much from the previously learned model(s)~\citep{kirkpatrick2017overcoming,chaudhry2018riemannian,schwarz2018progress,aljundi2019task}. 3) Rehearsal methods save exemplar raw data, called episodic memory, from previously learned tasks. When learning new tasks, these methods simultaneously learn on the new task and rehearse on episodic memories to retain past knowledge~\citep{chaudhry2019tiny,lopez2017gradient,chaudhry2018efficient,buzzega2020dark}. Other than saving the raw data points, pseudo-rehearsal like training a generative model to replay past experience is also a popular approach~\citep{shin2017continual}. For a comprehensive survey of existing continual learning methods, we refer the reader to tow survey papers~\citep{van2019three,delange2021continual}. In contrast to existing CL methods that focus on reducing forgetting, the CLPU problem requires the agent to deliberately forget a particular task upon request, while minimally influencing other knowledge. 

\paragraph{Machine Unlearning} Machine unlearning (MU) studies how to remove the effect of a specific training sample on a learning model per a user's request \citep{cao2015towards}. The most straightforward approach is to retrain the model on all data except the portion that has been removed, but this approach is in general impractical if the entire training set is large. To this end, typical MU approaches consider training multiple models on different shards of data so that unlearning only requires retraining a specific model on part of the dataset \citep{bourtoule2019machine}, or storing learned model parameters and their gradients for rapid retraining \citep{wu2020deltagrad}. There is also research focusing on MU with specific model or problem assumptions, such as linear models \citep{guo2019certified}, random forests \citep{brophy2021machine}, or k-means\citep{ginart2019making}.
%
Based on differential privacy, Golatkar et al. introduced ``scrubbing" that removes information from the weights of deep networks based on the Fisher Information Matrix \citep{golatkar2020eternal}. Mixed-Linear Forgetting proposes a tractable optimization problem by lineary approximating the amount of change in weights due to the addition of any training data \citep{golatkar2021mixed}. The above methods all consider the MU problem in general where the preserved dataset (e.g., data except the removed ones) is available, which is not the case in continual learning. Recently, a particularly relevant study first considers MU in the context of continual learning \citep{shibata2021learning}. However, their problem definition aims to make the model predict as wrongly as possible on the removed data, which does not in general protect the user's privacy. For instance, if the agent has learned task B that helps improve prediction on task A, which the agent is asked to forget, then completely random prediction on task A also reveals that the agent has learned on it before but asked to unlearn it later. In fact, as we observe from experimental results (Sec.~\ref{sec:experiment}), simply decreasing the model's performance on removed data is not private.
\section{Background}
In this section, we present the notation, definitions, and necessary background information to formalize CLPU. 

\subsection{Continual Learning}
\label{sec:bg-cl}
In continual learning (CL), an agent observes and learns $K$ tasks in a sequence. In this work, we assume each task $k \in [K]$\footnote{$[K]$ denotes $\{1,2,\dots,K\}$.} is a supervised learning task with a loss function $\ell^k: \mathcal{X} \times \mathcal{Y} \rightarrow \mathbb{R}$, a training dataset $D^k = \{(x^{k, \text{train}}_1, y^{k, \text{train}}_1), \dots, (x^{k, \text{train}}_n, y^{k, \text{train}}_n)\}$ and a testing dataset $D^k_\text{test} = \{(x^{k, \text{test}}_1, y^{k, \text{test}}_1), \dots, (x^{k, \text{test}}_{n'}, y^{k, \text{test}}_{n'})\}$. Here, $(x, y)$ are the raw data and labels where $x \in \mathcal{X}$ and $y \in \mathcal{Y}$. Assume the agent adopts a model $f_\theta$ parameterized by $\theta \in \mathbb{R}^d$. For instance, for a classification task, $\text{softmax}(f_\theta)$ produces a probability distribution over $\mathcal{Y}$. Then $\ell^k$ evaluates how well $f_\theta$ predicts $y$ given $x$. For instance, $\ell^k$ can be the standard cross-entropy loss for classification tasks (e.g., $\ell^k(x, y, f_\theta) = \log \sum_{y'} \exp \big(f_\theta(x)[y']\big) - f_\theta(x)[y]$). 

On learning task $k$, the agent loses its access to $D^{<k} = \{D^1, \dots, D^{k-1}\}$. After learning all $K$ tasks, the agent's objective is to achieve low loss on all test datasets $\{D^k_\text{test}\}_{k \in [K]}$. Assume the agent learns with a model $f$ that is parameterized by $\theta \in \mathbb{R}^m$. Denote the agent's model after learning task $k$ as $f_{\theta^{k}}$. Then, the overarching objective of a CL agent is to optimize
\begin{equation}
    \min_{\theta^K} \frac{1}{K}\sum_{k \in [K]} \mathbb{E}_{(x,y) \sim D^k_\text{test}} \bigg[ \ell^k(x, y, f_{\theta^K})\bigg].
    \label{eq:true-cl-obj}
\end{equation}
However, \eqref{eq:true-cl-obj} is hard to directly optimize using gradient-based methods (e.g., Stochastic Gradient Descent (SGD)) because of the dependency of $\theta^K$ on all previous $\theta^{k}$ such that $k < K$. Therefore, alternatively, from an induction point of view, the objective can be decomposed into $K$ objectives throughout the learning process. For learning the first task, the agent just optimizes the training loss $\ell^1$ on the training dataset $D^1$ as in standard supervised learning. For any task $k > 1$, the agent is asked to achieve low loss on task $k$ while maintaining its performance on previously learned tasks~\citep{lopez2017gradient}: 
\begin{equation}
    \min_{\theta \in \mathbb{R}^m} ~ \underbrace{\mathbb{E}_{(x,y) \sim D^k}\bigg[ \ell^k(x, y, f_\theta)\bigg]}_{\text{performance on task}~k}~~~\text{s.t.}~~~ \forall \tau < k,~~\underbrace{\mathbb{E}_{(x,y) \sim D^\tau} \bigg[ \ell^\tau(x, y, f_\theta) - \ell^\tau(x, y, f_{\theta^{(k-1)}})\bigg]}_{\text{forgetting on task}~\tau} \leq 0.
    \label{eq:cl-objective}
\end{equation}
Note that here we also replace the testing loss by training loss as the agent is not assumed to have access to test data during training. In practice, when the underlying $K$ tasks share the same loss function (e.g., they are all image classification tasks), which we assume for the rest of this paper, we can elide the superscript $k$ in $\ell^k$.

The main challenge for solving CL results from losing access to $D^{<k}$. Regularization-based methods assume all information learned from $D^{<k}$ is in $\theta^{k-1}$.  Thus they aim to minimize the training loss on $D^k$ while ensuring that $\theta$ stays close to $\theta^{k-1}$:
\begin{equation}
    \min_{\theta \in \mathbb{R}^m} ~ \mathbb{E}_{(x,y) \sim D^k}\bigg[ \ell(x, y, f_\theta)\bigg] + \alpha D(\theta, \theta^{k-1}),
\end{equation}
where $\alpha$ is a hyperparameter determining the strength of regularization and $D(\theta, \theta^{k-1}$) is a divergence measure that captures how similar $\theta$ is to $\theta^{k-1}$. Another popular and empirically more effective approach in CL is the rehearsal-based approach. These methods allow the agent to store a small number of exemplar data points, known as the \emph{episodic memory} $\{B^\tau\}_{\tau < k}$, for maintaining the agent's learned knowledge on previous tasks. In particular, $B^\tau$ stores $b^\tau \ll |D^\tau|$ i.i.d.~ sampled data points from $D^\tau$, optionally with $f_{\theta^\tau}$'s final layer output (a.k.a. the logits):
$$
B^\tau = \bigg\{ \big(x_i, y_i, h_i =f_{\theta^\tau}(x_i)\big) ~~\bigg|~~ (x_i, y_i) \stackrel{\text{i.i.d}}{\sim} D^\tau \bigg\}_{1 \leq i \leq b^\tau}.
$$
As a result, the general objective of rehearsal-based methods is
\begin{equation}
     \min_{\theta \in \mathbb{R}^m} ~ \frac{\beta}{|D^k|} \sum_{(x,y) \sim D^k} \ell(x, y, f_\theta) + (1-\beta) \sum_{\tau <k}\frac{1}{|B^\tau|}\sum_{(x',y',h') \sim B^\tau} \hat{\ell}(x', y', h', f_\theta),
    \label{eq:er-objective}
\end{equation}
where $\beta > 0$ is a hyper-parameter that weights the trade-off between learning new and preserving old knowledge. $\hat{\ell}$ can be the standard cross-entropy loss $\ell$, the knowledge distillation loss $\ell_\text{distill}(x, h, f_\theta) =  \KL\big({\softmax(f_\theta(x)) \mid\mid h}\big)$, or the mean-square-error between the predicted and saved logits $\ell_\text{mse}(x, h, f_\theta) = \frac{1}{2}||f_\theta(x) - h||^2_2$, which has shown strong performance at preserving past knowledge~\citep{buzzega2020dark}. Specifically, for the Dark Experience Replay++ (DER++) method ~\citep{buzzega2020dark}, $\hat{\ell}$ is a linear combination of the standard cross entropy loss and mean-square-error loss: $\hat{\ell}(x', y', h', f_\theta) = \alpha_1 \ell(x', y', f_\theta) + \alpha_2 \ell_\text{mse}(x', h', f_\theta)$.

\subsection{Machine Unlearning}
Unlike continual learning which studies learning over sequential tasks, contemporary research in machine unlearning (MU) mainly focuses on single-task learning~\citep{cao2015towards}. In particular, MU is often studied in the context of supervised learning, though extensions to other types of learning are relatively straightforward. Under the standard supervised learning setting, the agent is given a training dataset $D = \{(x_1, y_1), (x_2, y_2), \dots, (x_n, y_n)\}$. The agent applies a (stochastic) learning algorithm $A$ on $D$ to learn a model $f_\theta$ parameterized by $\theta$, such that $f_\theta$ achieves low empirical loss (e.g., $\frac{1}{n} \sum_{i=1}^n \ell(x_i, y_i, f_\theta)$ is small). Denote $A(D)$ as the distribution over the resulting model parameters $\theta$ when $A$ is applied on $D$.

A user can then request that the agent unlearn part of the dataset, which we call the forget set, $D_f \subset D$. Denote $D_r = D \setminus D_f$ as the retained dataset. Machine unlearning (MU) aims to find an unlearning algorithm $R_A$ that returns a model $\theta \sim R_A(D, A(D), D_f)$, which possesses no information about $D_f$ while performing well on $D_r$. In general, it is usually assumed that $|D_f| \ll |D|$, otherwise one can directly retrain a model on $D_r$. If the unlearned model from $R_A(D, A(D), D_f)$ has no information about $D_f$, an adversary cannot differentiate the model after unlearning from a model that is retrained on $D_r$, and we say that $(A, R_A)$ achieves \emph{exact unlearning}. The formal definitions are as follows.

\begin{mydef}[Exact Unlearning]
A pair of learning and unlearning algorithms $(A, R_A)$ achieve exact unlearning if
\begin{equation}
    \forall D,~D_f \subset D,~~A(D_r) =_d R_A(D, A(D), D_f),~~\text{where}~~ D_r = D \setminus D_f.
\end{equation}
Here $X =_d Y$ means $X$ and $Y$ share the same distribution.
\label{def:exact-unlearning}
\end{mydef}
The definition of exact unlearning is quite restrictive and therefore can be hard to achieve in practice. As a relaxation, \cite{ginart2019making} proposed the following definition of \emph{approximate unlearning}.
\begin{mydef}[Approximate $\delta$-Unlearning]
$(A, R_A)$ satisfies $\delta$-unlearning if
\begin{equation}
    \forall D,~D_f \subset D,~\text{and}~E \subseteq \mathbb{R}^d,~~P\big(R_A(D, A(D), D_f) \in E\big) \leq \delta^{-1} P\big(A(D_r) \in E\big).
\end{equation}
\label{def:eps-delta-unlearning}
\end{mydef}
These definitions are based on the assumption that the adversary can directly access the model parameters $\theta$ and therefore the definitions are based on the distribution over $\theta$. A more general assumption is that the adversary can only access the model via an output function $O(\theta, D)$, where $O: \Theta \times \mathcal{X} \longrightarrow \mathcal{O}$ and $\mathcal{X}$ denotes the space of input data. For instance, if $O(\theta, x) = f_\theta(x)$, it means the adversary only has access to the agent's prediction on any data point $x$. In that case, we can modify the definitions by replacing $R_A(\cdot)$ by $O \circ R_A(\cdot)$ and $A(\cdot)$ by $O \circ A(\cdot)$.

\paragraph{Remarks}
\begin{itemize}
    \item \textbf{MU depends on both $A$ and $R_A$.} As $R_A$ highly depends on the learning algorithm $A$, MU focuses on the design of both. But note that purely achieving exact unlearning without considering the model's performance is meaningless. For example, one can achieve exact unlearning trivially if $A$ yields a constant mapping and $R_A$ is an identity mapping. Therefore, the challenge is to maintain $R_A(D, A(D), D_f)$'s performance on $D_r$, while unlearning exactly.
    \item \textbf{MU differs from differential privacy (DP):} $\epsilon$-DP does not divide the data into $D_f$ and $D_r$ and requires that \emph{no} individual data point can significantly influence the model's prediction. But in MU (with exact unlearning), it is required that any data point $x \in D_f$ has \emph{zero} influence on the model's prediction after the unlearning, with no restrictions on the effects of data $x \in D_r$.
    \item \textbf{$\delta$-Unlearning is asymmetric} The above definitions implicitly assume that $P(\theta \mid A(D_r)) = 0 \Longrightarrow P(\theta \mid R_A(D, A(D), D_f)) = 0$. However, we do not assume the converse, meaning that it is permissible for $R_A$ to not generate models that could have been generated by $A(D_r)$.
\end{itemize}

\section{Problem and Method}
We start this section with a formal introduction to the continual learning and private unlearning (CLPU) problem. Then we introduce a straightforward solution to CLPU that achieves exact unlearning by saving extra models, thus sacrificing some space complexity compared to using a fixed-sized model throughout learning.

\subsection{CLPU: The Continual Learning and Private Unlearning problem}
In Continual Learning and Private Unlearning (CLPU), an agent receives $T$ requests from the user sequentially and is asked to learn from a pool of $K$ tasks in total. The $t$-th request $\RRR^t$ is a tuple $\RRR^t = (I^t, D^t, \rho^t)$. Here, $I^t \in [K]$ is the task ID, indicating the current task of interest. $D^t$ is either the training dataset $\{(x_i, y_i)\}_{i=1}^{|D^t|}$ or an empty set $\emptyset$ depending on what $\rho^t$ is. $\rho^t \in \{\mathbf{R}, \mathbf{T}, \mathbf{F}\}$ is a learning instruction:
\begin{itemize}
    \item $\rho^t = \mathbf{R}$: the user asks the agent to learn on task $i$ permanently.
    \item $\rho^t = \mathbf{T}$: the user asks the agent to temporarily learn on task $i$, which can be forgot in the future.
    \item $\rho^t = \mathbf{F}$: the user asks the agent to forget task $i$ with exact unlearning.
\end{itemize}
The agent keeps a dictionary $\Psi^t$ of the learned tasks' statuses, which, given $\RRR^t$, is updated by:
\begin{equation}
    \begin{cases}
    \Psi^t[I^t] 
    = (D^t, \rho^t) & \text{if}~~ \rho^t \in \{\mathbf{R}, \mathbf{T}\} \\
    \Psi^t \leftarrow \Psi^{(t-1)} \setminus \{I^t\} & \text{if}~~ \rho^t = \mathbf{F}.
    \end{cases}
    \label{eq:update-dictionary}
\end{equation}
Here, $\Psi \setminus I$ indicates the removal of the key $I$ as well as its corresponding values $(D, \rho)$ from $\Psi$.  If $I^t \in \Psi^{(t-1)}$ and $\rho^t = \mathbf{R}$, the agent is to fully memorize a task that has previously been temporarily learned with instruction $\mathbf{T}$. In both this case and the case when $\rho^t = \mathbf{F}$, we assume $D^t = \emptyset$ as there is no need for the user to provide the dataset for the same task twice. 

\begin{figure}[t]
    \centering
    \includegraphics[width=0.7\textwidth]{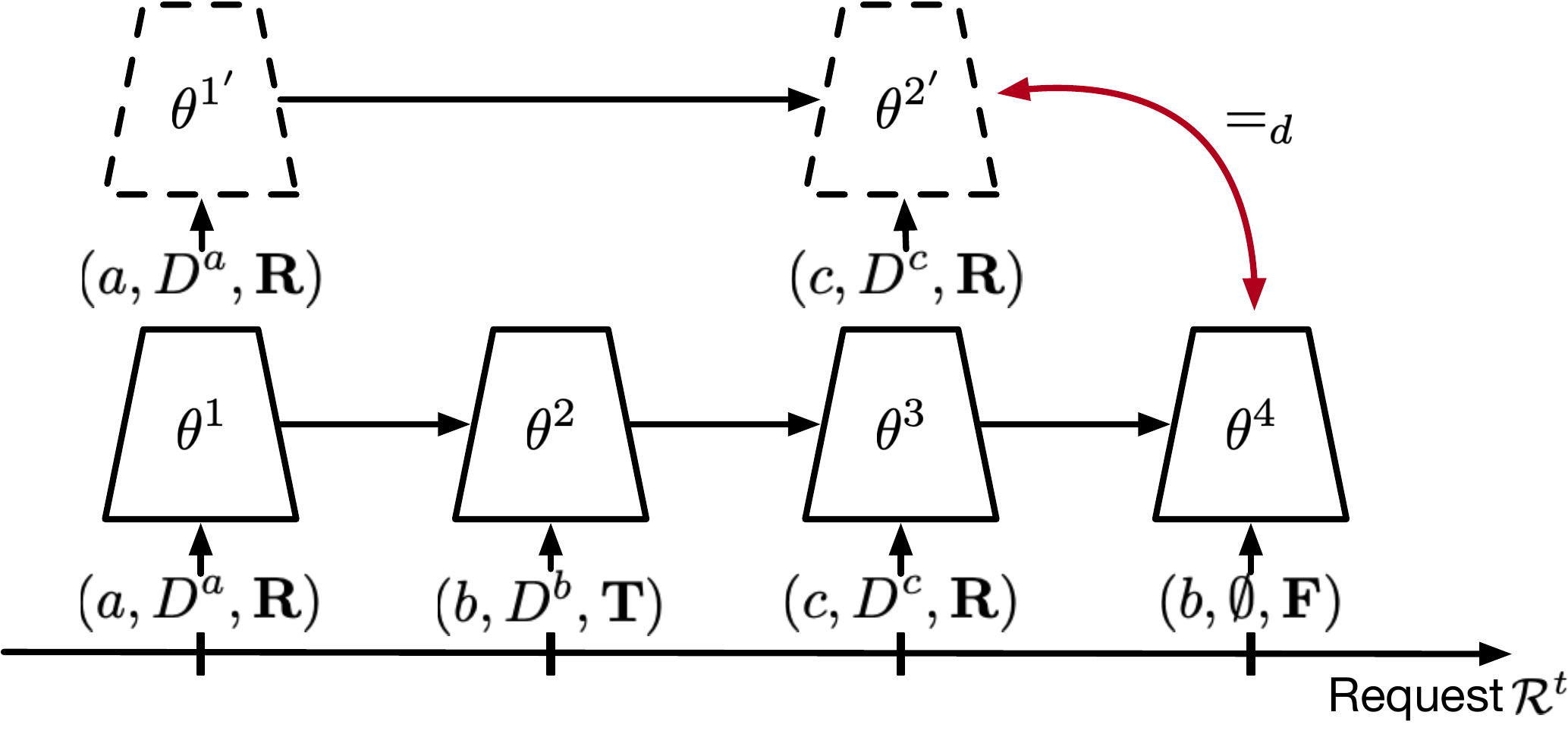}
    \caption{An illustration of the Continual Learning and private unlearning (CLPU) problem setting. After the agent has temporarily learned on task $b$ with data $D^b$, if the agent is later requested to unlearn task $b$, the unlearned model parameters $\theta^4$ should be indistinguishable from $\theta^{2'}$ in distribution as if the agent has never learned on task $b$. Except for the unlearn requests, the agent should perform continual learning over the remaining sequence of tasks.}
    \vspace{-10pt}
    \label{fig:clpu}
\end{figure}

Now we are ready to present the formulation of CLPU. Denote all requests up to the $(t-1)$-th request as $\RRR^{< t} = [\RRR^1, \RRR^2, \dots, \RRR^{(t-1)}]$.\footnote{We use list notation $[\RRR^1, \RRR^2, \dots]$ to indicate that the ordering matters.} A CLPU solution consists of a continual learning algorithm $A$ and an unlearning algorithm $R_A$. Let $f_\theta$ be the learning model parameterized by $\theta \in \mathbb{R}^m$ and $\theta^t$ denote the model parameter after processing the $t$-th request. Then both $A$ and $R_A$ map the previous model parameters and the current request to updated model parameters. In particular, we have the following recursion:
\begin{equation}
    \begin{cases}
        \theta^t \sim A(\theta^{(t-1)}, \RRR^t = (I^t, D^t, \rho^t)) & \text{if}~\rho^t \in \{\mathbf{R, T}\}, \\
        \theta^t \sim R_A(\theta^{(t-1)}, \RRR^t = (I^t, D^t, \rho^t)) & \text{if}~\rho^t = \mathbf{F}.
    \end{cases}
\end{equation}
For simplicity of notation, if all requests from $\RRR^s$ to $\RRR^t$ ($\RRR^{s:t}$ for short), have $\rho \in \{\mathbf{R}, \mathbf{T}\}$, then we denote $\theta^t \sim A(\theta^{s-1}, \RRR^{s:t})$. Additionally, denote $[\tau \in \Psi^{(t-1)}]$ as all tasks with $\rho \in \{\mathbf{R, T}\}$ that have been observed and not removed up to the $(t-1)$-th request. The objective of a CLPU agent when processing $\RRR^t$ is for $A$ and $R_A$ to output $\theta^t$ with the following properties:
\begin{equation}
    \theta^t \sim \begin{cases}
        \begin{array}{l}
        \argmin_{\theta} \mathbb{E}_{(x,y) \sim D^t} \bigg[\ell(x,y, f_{\theta}\rebuttal{)}\bigg]~~~\text{s.t}~~~ \\
        \qquad~~\forall \tau \in \Psi^{(t-1)},~\mathbb{E}_{(x,y) \sim D^\tau} \bigg[ \ell(x, y, f_{\theta}) - \ell(x, y, f_{\theta^{(t-1)}})\bigg] \leq 0
        \end{array}
        & \text{if}~\rho^t \in \{\mathbf{R, T}\}, \\
        R_A(\theta^{(t-1)}, \RRR^t)~~\text{s.t.}~~ \mathcal{D}\bigg(R_A(\theta^{(t-1)}, \RRR^t)~~\big|\big|~~ A(\theta^0, \RRR^{[\tau \in \Psi^{(t-1)}]})\bigg) = 0 & \text{if}~\rho^t = \mathbf{F}.
    \end{cases}
\end{equation}
Here, $\mathcal{D}(A ~||~B)$ is a distance between distributions $A$ and $B$. In other words, in the first case when $\rho^t \in \{\mathbf{R}, \mathbf{T}\}$, the expected loss cannot get any worse for any previously learned (but not forgotten) task. In the second case when $\rho^t = \mathbf{F}$, then the unlearned model parameters cannot be distinguished from the model parameters learned over the sequence of non-forgetting tasks with the divergence $\mathcal{D}$. The CLPU problem setting is illustrated in Fig.~\ref{fig:clpu}.


\paragraph{How does CLPU differ from CL and MU?} 1) In CLPU, in addition to exhibiting knowledge transfer as in CL, the agent also needs to unlearn specific tasks while maintaining all knowledge unrelated to the forgotten tasks. 2) Unlike MU where generally the agent learns on i.i.d.\  samples from the entire dataset $D$, in CLPU the agent learns online over different tasks and hence the ordering of the sequence of tasks matters. In addition, CLPU does not in general assume the agent can keep all previous data, which makes the unlearning (or more specifically the retention of learned knowledge) more difficult.

\subsection{CLPU-DER++: An Initial CLPU Algorithm}
\label{sec:clpu-der++}
In this section, we present a straightforward method to CLPU, named CLPU-DER++ as it adapts the DER++ method~\citep{buzzega2020dark} to the CLPU problem. The CLPU-DER++ method achieves exact unlearning upon request and learns continually over a sequence of tasks otherwise.

Inspired by Sharded, Isolated, Sliced, and Aggregated training (SISA)~\citep{bourtoule2019machine}, for each task with $\rho=\mathbf{T}$, the CLPU-DER++ agent creates an isolated temporary network with parameters $\hat{\theta}$. Based on the subsequent learning instruction for the same task, the agent either removes this isolated model ($\mathbb{F}$) or merges it with the main model ($\mathbb{R}$).

Specifically, we assume the agent maintains a main model with parameters $\theta_\text{main}$ and a set $N$ of temporary models. In other words, $\theta = \{\theta_\text{main}\} \cup N$. Upon the $t$-th request $\RRR^t = (I^t, D^t, \rho^t)$, there are four possible cases. \textbf{1)} If $\rho^t = \mathbf{R}$ and $I^t \notin \Psi^{(t-1)}$, then this is the first time the agent has observed task $I^t$, so the agent then performs conventional CL using Dark Experience Replay++ (DER++)~\citep{buzzega2020dark} and updates the main model parameters to $\theta^t_\text{main}$. 
\textbf{2)} If $\rho^t = \mathbf{T}$, in order to benefit from prior learning experience, the agent initializes an isolated model with parameters $\hat{\theta}$ copied from $\theta^{(t-1)}_\text{main}$, then directly performs SGD update on $\hat{\theta}$ using the dataset $D^t$ and includes the updated network to $N$ (e.g., $N \leftarrow N \cup \hat{\theta}^{I^t}$). In both cases, the agent stores episodic memory $B^{I^t}$ (See Sec.~\ref{sec:bg-cl}) for the task $I^t$.
\textbf{3)} If $\rho^t = \mathbf{R}$ and $I^t \in \Psi^{(t-1)}$, this means the agent has previously learned on task $I^t$ with a temporary network $\hat{\theta}^{I^t}$. Then the agent merges the knowledge learned in $\hat{\theta}^{I^t}$ into $\theta^t_\text{main}$ to reduce space and encourage knowledge transfer. To do so, CLPU-DER++ performs knowledge distillation on the combined episodic memories from task $I^t$ and the rest of the previously fully remembered tasks in $\Psi^t$. \textbf{4)} If $\rho^t = \mathbf{F}$, we simply remove the temporary network $\hat{\theta}^{I^t}$ from $N$. The details of CLPU-DER++ are presented in Alg.~\ref{alg:clpu-derpp}.

\begin{algorithm*}[t]
    \caption{Continual Learning and private unlearning - Dark Experience Replay++ (CLPU-DER++)}
    \begin{algorithmic}[1] 
        \STATE \textbf{Input}: Initial main model parameters $\theta^0_\text{main}$ and temporary networks $N = \emptyset$, initial task status dictionary $\Psi^0 = \emptyset$, the total number of user requests $T$, and memory sizes $\{b^t\}_{t=1}^T$. 
        \FOR{$t = 1: T$}
            \STATE Receive request $\RRR^t = (I^t, D^t, \rho^t)$.
            \STATE Update $\Psi^t$ by 
                $$
    \begin{cases}
    \Psi^t[I^t] 
    = (D^t, \rho^t) & \text{if}~~ \rho^t \in \{\mathbf{R}, \mathbf{T}\} \\
    \Psi^t \leftarrow \Psi^{(t-1)} \setminus \{I^t\} & \text{if}~~ \rho^t = \mathbf{F}.
    \end{cases}
                $$
            \STATE \textbf{Case I:} ~~$\rho^t =\mathbf{R}~\textbf{and}~I^t \notin \Psi^{(t-1)}$
            \STATE \quad Perform $H$ steps of SGD from $\theta_\text{main}^{(t-1)}$ by optimizing:
                $$
                 \theta^{t}_\text{main} = \argmin_{\theta \in \mathbb{R}^m} ~ \frac{1}{|D^t|} \sum_{(x,y) \sim D^t} \ell(x, y, f_\theta) + \frac{1}{|\Psi^{(t-1)}|} \sum_{i \in \Psi^{(t-1)}}\frac{1}{|B^i|}\sum_{(x', h') \sim B^i}\ell_\text{mse}(x', h', f_\theta).
                $$
            \STATE \quad Build the episodic memory $B^{I^t}$:
            $B^{I^t} = \big\{ \big(x_i, y_i, f_{\theta^t_\text{main}}(x_i)\big) ~~\big|~~ (x_i, y_i) \stackrel{\text{i.i.d}}{\sim} D^t \big\}_{1 \leq i \leq b^t}$.
            
            \STATE \textbf{Case II:} ~~$\rho^t = \mathbf{T}$
            \STATE \quad Initialize $\hat{\theta}^{I^t}$ from $\theta^{(t-1)}_\text{main}$.
            \STATE \quad Perform $H$ steps of SGD on $\hat{\theta}^{I^t}$ by optimizing:
                $$
                 \hat{\theta}^{I^t} = \argmin_{\theta \in \mathbb{R}^m} ~ \frac{1}{|D^t|} \sum_{(x,y) \sim D^t} \ell(x, y, f_\theta) + \frac{1}{|\Psi^{(t-1)}|} \sum_{i \in \Psi^{(t-1)}}\frac{1}{|B^i|}\sum_{(x', h') \sim B^i}\ell_\text{mse}(x', h', f_\theta).
                $$
            \STATE \quad Store the temporary network: $N \leftarrow N \cup \{\hat{\theta}^{I^t}\}$. 
            \STATE \quad
            Build the episodic memory $B^{I^t}$:
            $B^{I^t} = \big\{ \big(x_i, y_i, f_{\theta^t_\text{main}}(x_i)\big) ~~\big|~~ (x_i, y_i) \stackrel{\text{i.i.d}}{\sim} D^t \big\}_{1 \leq i \leq b^t}$.
            
            \STATE \textbf{Case III:} ~~$\rho^t = \mathbf{R}$ and $I^t \in \Psi^{(t-1)}$
            \STATE \quad Merge $\hat{\theta}^{I^t}$ back to $\theta_\text{main}^t$ by performing $H$ step of SGD and optimize:
            \STATE 
            $$
             \theta^t = \argmin_{\theta \in \mathbb{R}^m} ~ \frac{1}{|B^{I^t}|} \sum_{(x,h) \sim B^{I^t}} \ell_\text{mse}(x, h, f_\theta) + \frac{1}{|\Psi^t|} \sum_{i \in \Psi^t}\frac{1}{|B^i|}\sum_{(x',h') \sim B^i} \ell_\text{mse}(x', h', f_\theta).
            $$
            \STATE \quad Remove the temporary network:  $N = N \setminus \{\hat{\theta}^{I^t}\}$.
            
            \STATE \textbf{Case IV:} ~~$\rho^t = \mathbf{F}$
            \STATE \quad Remove the temporary network:  $N = N \setminus \{\hat{\theta}^{I^t}\}$.
        \ENDFOR
    \end{algorithmic}
    \label{alg:clpu-derpp}
\end{algorithm*}

\paragraph{Remark} CLPU-DER++ achieves \emph{exact} unlearning (See Def.~\ref{def:exact-unlearning}) by construction.  For any task $k$ that the agent has learned previously and then attempts to unlearn, the unlearn process only involves removing the relevant temporary model from $N$ and the corresponding episodic memory: it does not influence the main model parameter $\theta_\text{main}$. On the other hand, CLPU-DER++ achieves privacy at the expense of memory, as it stores a full extra model for each temporary learning task, which can be particularly important for large modern neural architectures.

\section{Experimental Results}
\label{sec:experiment}

In this section, we first introduce the experiment setup and introduce how we form novel benchmarks by adapting conventional CL datasets for the CLPU problem. Then we introduce the evaluation metrics designed for measuring the agent's performance in terms of both continual learning and private unlearning. In the end, we present the evaluation results by comparing CLPU-DER++ against the following baseline methods: sequential learning (Seq), indepdent learning (Ind), Elastic Weight Consolidation (EWC)~\citep{kirkpatrick2017overcoming}, Learning without Forgetting (LwF)~\citep{li2017learning} , Experience Replay (ER)~\citep{chaudhry2019tiny}, Dark Experience Replay++ (DER++)~\citep{buzzega2020dark}, and Learning with Selective Forgetting (LSF)~\citep{shibata2021learning}. All the above baselines except LSF are state-of-the-art CL methods, but we adapt some of them for the CLPU setting. In particular, for sequential learning, the agent performs SGD directly over the sequence of tasks. For independent learning, the agent creates a new model for each new task, and removes a model if the user requests to unlearn the corresponding task. For ER and DER++, for an unlearning task, we remove the corresponding episodic memory and let the agent perform normal ER and DER++ updates on the remaining episodic memories and predict uniform distributions for the forgotten task to accelerate forgetting.

\subsection{CLPU Experiment Setup} 
We consider four conventional CL benchmarks: 
rotation MNIST (rot-MNIST), permutation MNIST (perm-MNIST), split CIFAR-10 and split CIFAR-100. rot-MNIST and perm-MNIST datasets are formed by rotating the images and randomly permuting the pixels of the images, respectively, in the MNIST dataset. Each task is a 10-class classification task. Split CIFAR-10 and split CIFAR-100 are formed by treating the 10 classes in CIFAR-10 as five 2-class classification tasks, and the 100 classes in CIFAR-100 as five 20-class classification tasks. To be consistent, we build rot-MNIST and perm-MNIST also with 5 sequential tasks. Then, to adapt these datasets to the CLPU setting, we form the following sequence of requests:
$$\RRR^{1:8} = \big[(1, D^1, \bm{R}),~~(2, D^2, \bm{T}),~~(3, D^3, \bm{T}),~~(4, D^4, \bm{R}),~~ (1, \emptyset, \bm{R}),~~ (2, \emptyset, \bm{F}),~~ (5, D^5, \bm{T}),~~ (5, \emptyset, \bm{F})\big].$$
The corresponding sequence of requests that involve no unlearning is therefore
$$
\hat{\RRR}^{1:4} = \big[(1, D^1, \bm{R}),~~(3, D^3, \bm{T}),~~(4, D^4, \bm{R}),~~ (1, \emptyset, \bm{R})\big].
$$


For all datasets, we use the SGD optimizer without momentum with $0.0005$ weight decay. For all datasets, the learning rate is set to $0.01$ and we perform 10 epochs of training for each task. When the agent is asked to unlearn a task, we also perform 10 epochs of the algorithm-specific unlearn updates. The implementations of the baseline methods are adapted from the open-source DER implementation.\footnote{DER code from \url{https://github.com/aimagelab/mammoth}.}

\subsection{Evaluation Metrics}
To evaluate a method on CLPU, we consider metrics both for continual learning and for private unlearning. To measure the method's performance on continual learning, we report the final average accuracy (ACC) of the model over all tasks that remain in the final task status dictionary $\Phi$, as well as the forgetting measure (FM), which is the average drop in performance on each task, compared to the model's performance when the agent first learned these tasks. Note that all evaluations are done on holdout testing data $\{D^k_\text{test}\}$ for each task $k$. To be specific, denote $a^t_s$ as the agent's prediction accuracy on task $s$'s test dataset $D^s_\text{test}$ after processing user's $t$-th request, then we define
\begin{equation}
    \text{ACC} = \sum_{t=1}^T \sum_{s \in \Phi^t} a^t_s~~~~\text{and}~~~~\text{FM} = \sum_{t=1}^T \sum_{s \in \Phi^t} a^{\tau(s)}_s - a^t_s,~~~~\text{where}~~\tau(s) = \argmin_t~~(I^t = s).
\end{equation}
In short, ACC measures how well the agent performs on the tasks with $\rho \in \{\mathbf{R}, \mathbf{T}\}$ after processing all requests. FM measures how much the agent forgets on the same set of tasks compared to when they were first learned.

In addition to the above two metrics for evaluating the continual learning performance, we also compare, on all tasks with $\rho = \mathbf{F}$, the divergence between the model's output distribution on that task after unlearning versus the distribution that would have resulted had the agent not learned the task. In other words, we use the output function $O(\theta, x) = f_\theta(x)$ because comparing the distributions of $f_\theta(\cdot)$ with different $\theta$s is more computationally efficient than directly comparing the distributions of different $\theta$s.
Concretely, for all requests $\RRR^t$ such that $\rho^t = \mathbf{F}$, we measure how different $f_{\theta^t}$ is from $f_{\theta^{t'}}$, where $\theta^t \sim A(\theta^0, \RRR^{\leq t})$ and $\theta^{t'} \sim A(\theta^0, \RRR^{[\tau \in \Psi^{t}]})$. To measure the difference, we train $c$ models using different random seeds on $\RRR^{\leq t}$ to get parameters $\{\theta^t_1, \dots, \theta^t_c\}$, and similarly get $c$ model parameters $\{\theta^{t'}_1, \dots, \theta^{t'}_c\}$ by training on $ \RRR^{[\tau \in \Psi^{t}]}$. After that, for each pair of models that are trained on $\RRR^{[\tau \in \Psi^{t}]}$, we calculate the in-group Jensen-Shannon (IJSD) distance between their outputs on the testing dataset $D^{I^t}_\text{test}$.\footnote{We use Jensen-Shannon Distance because it is a symmetric divergence for comparing probability distributions.} Similarly, for any model trained on $\RRR^{[\tau \in \Psi^{t}]}$ and any other model trained on $\RRR^{\leq t}$, we also calculate their output distributions' Jenson-Shannon distance, which we call the Across-group Jensen-Shannon Distance (AJSD). For the entire sequence of requests, we average over the number of unlearning tasks for both IJSD and AJSD. Therefore, in total we have $\frac{c(c-1)}{2}$ distances for IJSD and $c^2$ distances for AJSD, over the entire sequence of requests. Formally,
\begin{equation}
    \begin{split}
    \text{IJSD} &= \bigg\{
    \sum_{1 \leq i < j \leq c} \frac{1}{|\sum_{t\in[T], \rho^t=\mathbf{F}}1|}\sum_{t \in [T], \rho^t = \mathbf{F}} \mathbb{E}_{(x,y) \sim D^{I^t}}\bigg[\JSD\bigg(f(x ; \theta^{t'}_i) ~\bigg|\bigg|~ f(x; \theta^{t'}_j)\bigg)\bigg]\bigg\}, \\
    \text{AJSD} &= \bigg\{
    \sum_{i, j \in [c]} \frac{1}{|\sum_{t\in[T], \rho^t=\mathbf{F}}1|}\sum_{t \in [T], \rho^t = \mathbf{F}} \mathbb{E}_{(x,y) \sim D^{I^t}}\bigg[\JSD\bigg(f(x ; \theta^{t}_i) ~\bigg|\bigg|~ f(x; \theta^{t'}_j)\bigg)\bigg]\bigg\}. \\
    \end{split}
\end{equation}

After the IJSDs and AJSDs are calculated, we measure the ratio of the absolute difference between the average of IJSD and AJSD over the average of IJSD, which we call JS-ratio, and the proportion of AJSD that are smaller than the maximum of IJSD, which we call the In-Range Rate (IRR). Formally, 
\begin{equation}
    \text{JS-ratio} = \frac{\big| \frac{1}{|\text{IJSD}|} \sum_{d \in \text{IJSD}} d - \frac{1}{|\text{AJSD}|} \sum_{d \in \text{AJSD}} d\big|}{\frac{1}{|\text{IJSD}|} \sum_{d \in \text{IJSD}} d },~~~~\text{and}~~~~ \text{IRR} = \frac{\sum_{d \in \text{AJSD}} \mathds{1}\big(d \leq \max(\text{IJSD})\big) }{\big| \text{AJSD}\big|}.
\end{equation}

\begin{table}[t!]
    \centering
    \begin{tabular}{lcccccc}
    \toprule
    & \multicolumn{6}{c}{Perm-MNIST}\\
    \cmidrule(lr){2-7}
    Method & ACC($\uparrow$) & FM($\downarrow$) & IJSD & AJSD & JS-ratio($\downarrow$) & IRR($\uparrow$) \\
    \midrule
Ind (Upper Bound)                        & 95.59 \fs { 0.05 } & 0.00 \fs { 0.00 } & 0.01 \fs { 0.00 }  & 0.01 \fs { 0.00 } & 0.14 & 0.96 \\
\cmidrule(lr){2-7}
Seq                                                & 75.75 \fs { 2.44 } & 19.97 \fs { 2.45 } & 0.17 \fs { 0.05 }  & 0.92 \fs { 0.03 } & 4.47 & 0.00 \\
EWC                                                & 93.67 \fs { 0.25 } & 0.45 \fs { 0.18 } & 0.04 \fs { 0.01 }  & 0.65 \fs { 0.01 } & 13.73 & 0.00 \\
ER                                                 & 91.83 \fs { 0.25 } & 3.96 \fs { 0.24 } & 0.11 \fs { 0.02 }  & 0.73 \fs { 0.01 } & 5.92 & 0.00 \\
LwF                                                & 79.09 \fs { 3.19 } & 17.12 \fs { 3.23 } & 0.08 \fs { 0.02 }  & 0.83 \fs { 0.03 } & 9.41 & 0.00 \\
LSF                                                & 91.18 \fs { 0.24 } & \textbf{0.44} \fs { 0.08 } & 0.06 \fs { 0.01 }  & 0.49 \fs { 0.02 } & 7.49 & 0.00 \\
DER++                                              & \textbf{93.88} \fs { 0.14 } & 2.01 \fs { 0.14 } & 0.07 \fs { 0.01 }  & 0.66 \fs { 0.01 } & 8.82 & 0.00 \\
CLPU-DER++ (scratch)                               & 93.26 \fs { 0.25 } & 2.33 \fs { 0.21 } & 0.10 \fs { 0.01 }  & 0.09 \fs { 0.02 } & \textbf{0.09} & \textbf{1.00} \\
CLPU-DER++                                         & 93.48 \fs { 0.25 } & 2.25 \fs { 0.30 } & 0.10 \fs { 0.01 }  & 0.09 \fs { 0.02 } & 0.13 & 0.96 \\
    \midrule
    & \multicolumn{6}{c}{Rot-MNIST}\\
    \cmidrule(lr){2-7}
    Method & ACC($\uparrow$) & FM($\downarrow$) & IJSD & AJSD & JS-ratio($\downarrow$) & IRR($\uparrow$) \\
    \midrule
    
Ind (Upper Bound)                        & 95.53 \fs { 0.06 } & 0.00 \fs { 0.00 } & 0.01 \fs { 0.00 }  & 0.01 \fs { 0.00 } & 0.14 & 0.96 \\
\cmidrule(lr){2-7}
Seq                                                & 90.88 \fs { 0.43 } & 5.66 \fs { 0.38 } & 0.14 \fs { 0.02 }  & 0.80 \fs { 0.02 } & 4.88 & 0.00 \\
EWC                                                & 94.75 \fs { 0.12 } & \textbf{0.29} \fs { 0.12 } & 0.09 \fs { 0.01 }  & 0.72 \fs { 0.01 } & 7.50 & 0.00 \\
ER                                                 & 95.12 \fs { 0.18 } & 1.39 \fs { 0.17 } & 0.16 \fs { 0.02 }  & 0.79 \fs { 0.02 } & 3.87 & 0.00 \\
LwF                                                & 95.72 \fs { 0.19 } & 0.87 \fs { 0.18 } & 0.07 \fs { 0.01 }  & 0.76 \fs { 0.01 } & 9.60 & 0.00 \\
LSF                                                & 92.56 \fs { 0.09 } & 0.30 \fs { 0.07 } & 0.08 \fs { 0.01 }  & 0.65 \fs { 0.02 } & 6.94 & 0.00 \\
DER++                                              & \textbf{95.94} \fs { 0.09 } & 0.36 \fs { 0.08 } & 0.12 \fs { 0.02 }  & 0.74 \fs { 0.02 } & 5.38 & 0.00 \\
CLPU-DER++ (scratch)                               & 94.69 \fs { 0.11 } & 1.02 \fs { 0.10 } & 0.13 \fs { 0.02 }  & 0.11 \fs { 0.04 } & \textbf{0.14} & \textbf{1.00} \\
CLPU-DER++                                         & 95.37 \fs { 0.12 } & 0.91 \fs { 0.09 } & 0.14 \fs { 0.02 }  & 0.11 \fs { 0.04 } & 0.17 & \textbf{1.00} \\
    \bottomrule 
    \end{tabular}
    \caption{Performance of CLPU-DER++ against baseline methods on the Perm-MNIST and Rot-MNIST CLPU benchmarks. We report the mean and standard deviation for each result over 5 independent runs. The best results for each metric are bolded. }
    \label{tab:clpu-mnist}
\end{table}

\begin{table}[t!]
    \centering
    \begin{tabular}{lcccccc}
    \toprule
    & \multicolumn{6}{c}{Split-CIFAR10}\\
    \cmidrule(lr){2-7}
    Method & ACC($\uparrow$) & FM($\downarrow$) & IJSD & AJSD & JS-ratio($\downarrow$) & IRR($\uparrow$) \\
    \midrule

Ind (Upper Bound)                        & 91.84 \fs { 0.94 } & 0.00 \fs { 0.00 } & 0.03 \fs { 0.02 }  & 0.03 \fs { 0.01 } & 0.12 & 1.00 \\
\cmidrule(lr){2-7}
Seq                                                & 76.73 \fs { 2.25 } & 15.38 \fs { 2.67 } & 0.09 \fs { 0.03 }  & 0.18 \fs { 0.02 } & 0.95 & 0.04 \\
EWC                                                & 79.43 \fs { 1.51 } & 11.83 \fs { 2.08 } & 0.14 \fs { 0.05 }  & 0.20 \fs { 0.03 } & 0.40 & 0.76 \\
ER                                                 & 90.44 \fs { 0.57 } & 0.83 \fs { 3.36 } & 0.07 \fs { 0.02 }  & 0.16 \fs { 0.01 } & 1.07 & 0.00 \\
LwF                                                & 88.18 \fs { 1.87 } & 4.34 \fs { 2.34 } & 0.06 \fs { 0.02 }  & 0.16 \fs { 0.03 } & 1.68 & 0.00 \\
LSF                                                & 90.00 \fs { 2.30 } & 2.48 \fs { 2.05 } & 0.07 \fs { 0.02 }  & 0.14 \fs { 0.03 } & 0.90 & 0.48 \\
DER++                                              & \textbf{91.26} \fs { 0.63 } & \textbf{0.73} \fs { 3.54 } & 0.03 \fs { 0.01 }  & 0.07 \fs { 0.01 } & 1.11 & 0.60 \\
CLPU-DER++ (scratch)                       & 89.52 \fs { 1.46 } & 1.14 \fs { 2.17 } & 0.03 \fs { 0.01 }  & 0.03 \fs { 0.02 } & 0.04 & \textbf{0.92} \\
CLPU-DER++                                         & 90.12 \fs { 1.65 } & 1.89 \fs { 2.20 } & 0.03 \fs { 0.01 }  & 0.03 \fs { 0.01 } & \textbf{0.00} & \textbf{0.92} \\
    \midrule 
    & \multicolumn{6}{c}{Split-CIFAR100}\\
    \cmidrule(lr){2-7}
    Method & ACC($\uparrow$) & FM($\downarrow$) & IJSD & AJSD & JS-ratio($\downarrow$) & IRR($\uparrow$) \\
    \midrule

Ind (Upper Bound)                        & 63.86 \fs { 0.55 } & 0.00 \fs { 0.00 } & 0.17 \fs { 0.01 }  & 0.17 \fs { 0.01 } & 0.00 & 0.96 \\
\cmidrule(lr){2-7}
Seq                                                & 44.34 \fs { 0.84 } & 24.36 \fs { 2.44 } & 0.44 \fs { 0.03 }  & 1.09 \fs { 0.03 } & 1.47 & 0.00 \\
EWC                                                & 45.39 \fs { 1.74 } & 20.08 \fs { 1.42 } & 0.63 \fs { 0.03 }  & 1.27 \fs { 0.04 } & 1.02 & 0.00 \\
ER                                                 & 61.66 \fs { 1.27 } & 7.69 \fs { 1.68 } & 0.51 \fs { 0.03 }  & 1.11 \fs { 0.03 } & 1.18 & 0.00 \\
LwF                                                & 61.25 \fs { 2.73 } & 8.60 \fs { 1.01 } & 0.39 \fs { 0.03 }  & 1.06 \fs { 0.03 } & 1.71 & 0.00 \\
LSF                                                & 37.92 \fs { 2.14 } & 26.88 \fs { 2.09 } & 0.70 \fs { 0.03 }  & 1.09 \fs { 0.05 } & 0.54 & 0.00 \\
DER++                                              & \textbf{66.66} \fs { 0.69 } & \textbf{2.84} \fs { 0.59 } & 0.31 \fs { 0.03 }  & 0.70 \fs { 0.02 } & 1.24 & 0.00 \\
CLPU-DER++ (scratch)                       & 61.51 \fs { 0.76 } & 3.46 \fs { 1.18 } & 0.21 \fs { 0.01 }  & 0.19 \fs { 0.03 } & \textbf{0.08} & \textbf{0.96} \\
CLPU-DER++                                         & 63.90 \fs { 0.77 } & 3.90 \fs { 1.05 } & 0.22 \fs { 0.01 }  & 0.21 \fs { 0.04 } & \textbf{0.08} & \textbf{0.96} \\
    \bottomrule 
    \end{tabular}
    \caption{Performance of CLPU-DER++ against baseline methods on on the Split-CIFAR10 and Split-CIFAR100 CLPU benchmarks. We report the mean and standard deviation for each result over 5 independent runs. The best results for each metric are bolded. }
    \label{tab:clpu-cifar}
\end{table}

\subsection{Results}
In Tab.~\ref{tab:clpu-mnist}-\ref{tab:clpu-cifar}, we report the comparison of CLPU-DER++ against the previously mentioned baseline methods on 4 benchmark datasetes. Specifically, we report the ACC, FM, JS-ratio and IRR metrics from previous section. To provide more information, we also report the mean and standard deviation of the sets IJSD and AJSD.\footnote{In Tab.~\ref{tab:clpu-mnist}-\ref{tab:clpu-cifar}, we abuse the notations of IJSD and AJSD a bit and directly report the mean and standard deviation under them.}

From the table, we can see that CLPU-DER++ achieves the best JS-ratio and IRR among all methods. In contrast, all baseline methods achieve high JS-ratio and very low IRR, meaning that the unlearning indeed reveals that the model
has learned on the unlearned task previously. On the other hand, in terms of the CL metrics, DER++  achieves the best CL performance with CLPU-DER++ finishing a close second. The  difference is due to the fact that when merging the temporary network back into the main model, the CLPU-DER++ agent essentially performs knowledge distillation to distill the knowledge from a temporarily learned task back to the main model. However, it is known that knowledge distillation often cannot fully recover the original model's performance.
Lastly, we observe that when creating a temporary network, initializing from the main model (line 9 of Alg.~\ref{alg:clpu-derpp}) results in better performance compared to initializing from scratch (CLPU-DER++ (scratch)).
\vspace{-2pt}
\section{Conclusion and Future Work}
\vspace{-2pt}
In this work, we propose a novel continual learning and private unlearning (CLPU) problem and provide its formal formulation. In addition, we introduce a straightforward but exact unlearning method to solve CLPU, as well as novel metrics and adapted benchmark problems to evaluate any CLPU methods. 
There are many interesting future directions for the CLPU problem. First, as shown in Fig.~\ref{fig:pareto}, CLPU-DER++ is an initial solutionn that achieves exact privacy and good knowledge transfer ability. It will be interesting to extend it to the $\delta$-unlearning setting while reducing the space complexity by saving fewer models. Second, it is important to understand theoretically what an optimal CLPU method can achieve. Note that in principle it might be impossible to reach the optima of the three objectives in Fig.~\ref{fig:pareto} simultaneously. Lastly, it is also interesting to study how the performance of any CLPU method can be affected by the relationship of different tasks. Intuitively, similar tasks should encourage better continual learning performance but make privately unlearning more difficult.
\vspace{-2pt}
\section{Acknowledgement}
\vspace{-2pt}
\thanks{This work has taken place in the Learning Agents Research
Group (LARG) at UT Austin.  LARG research is supported in part by NSF
(CPS-1739964, IIS-1724157, FAIN-2019844), ONR (N00014-18-2243), ARO
(W911NF-19-2-0333), DARPA, GM, Bosch, and UT Austin's
Good Systems grand challenge.  Peter Stone serves as the Executive
Director of Sony AI America and receives financial compensation for
this work.  The terms of this arrangement have been reviewed and
approved by the University of Texas at Austin in accordance with its
policy on objectivity in research.}


\newpage
\bibliography{collas2022_conference}
\bibliographystyle{collas2022_conference}


\end{document}